\documentclass[conference]{IEEEtran}
\IEEEoverridecommandlockouts
\usepackage{cite}
\usepackage{amsmath,amssymb,amsfonts}
\usepackage{algorithmic}
\usepackage{graphicx}
\usepackage{textcomp}
\usepackage{xcolor}
\def\BibTeX{{\rm B\kern-.05em{\sc i\kern-.025em b}\kern-.08em
    T\kern-.1667em\lower.7ex\hbox{E}\kern-.125emX}}
\begin{document}

\title{AI based Log Analyser: A Practical Approach\\
{\footnotesize \textsuperscript{ }}
\thanks{Identify applicable funding agency here. If none, delete this.}
}

\author{\IEEEauthorblockN{ Jonathan Pan}
\IEEEauthorblockA{\textit{Home Team Science and Technology Agency} \\
Singapore \\
Jonathan\textunderscore{}Pan@htx.gov.sg}

}

\maketitle

\begin{abstract}
The analysis of logs is a vital activity undertaken for fault or cyber incident detection, investigation and technical forensics analysis for system and cyber resilience. The potential application of AI algorithms for Log analysis could augment such complex and laborious tasks.  However, such solution has its constraints the heterogeneity of log sources and limited to no labels for training a classifier. When such labels become available, the need for the classifier to be updated. This practice-based research seeks to address these challenges with the use of Transformer construct to train a new model with only normal log entries. Log augmentation through multiple forms of perturbation is applied as a form of self-supervised training for feature learning. The model is further finetuned using a form of reinforcement learning with a limited set of label samples to mimic real-world situation with the availability of labels. The experimental results of our model construct show promise with comparative evaluation measurements paving the way for future practical applications. 
\end{abstract}


\section{Introduction}
The analysis of logs for anomalies is an important research topic with practical importance in the field of failure analysis [1], [2] and security threat detection [3], [4]. Logs are generated by systems or applications that are codified and configured to report relevant information about the state of the applications or software while running. Here, the application may refer to any software running to perform specific task or tasks. It could be mobile application, operating system running inside an Internet of Things (IoTs) device, a cloud compute node performing a computational task. It could also be an environment of compute nodes working collectively on multiple tasks.  Such logs and their log entries are generated based on its current configuration at the time of the log generation. These entries are also affected by the state of the application during its execution and its dependent factors that may originate from within the operating environment and executing platform of the application. It would be affected by external factors like users or external systems interacting with the application. 

These internal and external factors affecting the log generation may change abruptly and progressively over time resulting in corresponding log entries being included into the log generation process. These factors may originate from planned changes like planned maintenance tasks. They may also originate from unplanned activities. Additionally, these changes may be induced by benign or malicious intent. For the latter, with the intent to evade detection, even if the logs are not tampered, its entries will be elusive to classical detection techniques. These further complicates the composition of logs to be analyzed. 

The objective of performing analysis on logs is done to facilitate the detection of anomalous activities so that immediate or corresponding remediation may be done to contain or remediate the issue recorded in the logs. This is part of the attempt to enhance system resiliency against system faults, degradation and intentionally induced cyber physical attacks. It is also used to facilitate the investigation or analysis of what may have induced the occurrence of such anomalous activities. The scope of this research work is on the detection of such anomalous activities from the logs. 

There is many research work done to develop AI algorithms to detect anomalies from logs. However, log analysis using AI algorithms require extensive preparation [6][7]. The capability of log analysis using AI algorithms to detect anomalies has several challenges and constraints to deal with before it contributes significantly to its intended objectives of keeping system resilient. This is namely in dealing with the availability of anomalous data points or absence of labelled datasets that in turn facilitate the building of AI models to detect such occurrences. Also, the constraints with the task of gathering such labels to train the model due to inert limits of spotting or classifying anomalous patterns in logs. When such labels are available, they would be limited in numbers and frequency. Hence updating the model would be constrained too.  This practice based research work seeks to address these challenges through its contribution with the following which we believe that it provides a novel and practical approach to develop and upkeep an AI model for log analysis.

\begin{itemize}
\item The log analysis model uses a Transformer based classifier trained using self-supervised learning approach to learn the feature representation of normal log entries without any need for log preprocessing. Log augmentation is used through stochastic perturbation of log entries to train an anomaly classifier. Hence, a log analysis model can easily be created using normal log entries. 
\item The discovered labels to log entries can be used to update the model using a simple form of reinforcement learning. The model performs significantly better after it’s training update. Hence, the log analysis model can easily be updated and improve.
\end{itemize}

In the next section, we will cover the challenges and complexity of performing log analysis. This is followed by a review of current log analysis algorithms. A coverage of the algorithm that we have developed follows then with the details of an experimentation setup with its evaluation. This paper concludes with a conclusion and its future research direction.

\section{Background Information}

In this section, we articulate the background information related to the need for the analysis of logs, its complexity and challenges with current log analysis algorithms.

\subsection{Need for Analysis of Logs }

Logs are generated by software driven applications running on systems or devices to provide information to aid developers and system engineers with their analysis of system’s state and condition. It is also used as a form of audit trail to log the occurrences of events in chronological manner. The analysis of logs is also done to facilitate investigation after the occurrence of an incident related to the system that generates the logs. This incident could be in the form of system defect and a malicious or unintended breach of the system. With investigation, the log could provide the means to reconstruct the occurrence of the incident. With such forms of analysis, an investigator or system engineer would attempt to identify the occurrence of anomalous events through the logs. However, to identify such anomalies, one would need to know how to spot such anomalies from voluminous entries posted into the log files. 

\subsection{Log Analysis Challenges }

The form for logs is typically unique to how the software has been developed or configured to post entries into these textual files. Also, each system or software component may adopt its own logging format and information lexicon representation that details the state of the run-time system at the time of its log posting. Such information within the logs is contextual to the environment to which the system resides in. For example, information like the IP addresses or hostnames or resource identities. Entries in the logs are dependent also on the configuration surrounding the involved system and their own respective environmental conditions. In additional to the contextual settings, the log entries are sequenced by its chronological occurrence of events or state. Hence such log entries have a time dimension. 

Hence, the analysis of such log datasets requires contextual understanding of the system or component that generates such logs. Also, the analysis requires the means to classify or distinguish what is a normal log entry and what is not a normal log entry. For the latter, such information of recognizing an abnormal entry would be constrained to what may be conceivable based on the engineering design of the system involved or known instances of events that could cause an anomalous event like a cyber security breach attempt. However, there will be instances where such information or knowledge is only acquired through the occurrence of the event that in turn induces the anomalous log entries. Hence the challenges with log analysis are the need for contextual knowledge and limitedly available information about the form of anomalies that could occur.

\section{Related Work}
In this section, we review the current log analysis algorithmic development and their strengths and limitations.

\subsection{Multi-staged Log Processing}\label{AA}
The current log analysis algorithmic designs typically involve multiple stages of log processing before analysis is applied. It typically starts with log parsing that converts raw logs into structured data features. These extracted features would undergo further transformation as they are typically represented as textual features and would be converted to numerical forms. Log partitioning typically follows that involves converting the contiguous log into associative partitions to improve anomaly classification. This may involve the use of time-based partitions, partitions organized by windows of similar or compatible operations or identifier-based divisions of log entries. Finally, the anomaly detection algorithm would then be applied after these pre-processing.  

Thus far, there are very few developmental attempts to develop an integrated model that could ingest raw log data for immediate model training and inference. Based on our survey, one by Hashemi and Mäntylä [5] and Le and Zhang [16] ingress logs without log parsers. Le and Zhang observed that log parsers could cause inaccurate log parsing and not handle Out-of-vocabulary (OOV) words well. Our model applies its anomaly detection directly to the raw log entries.

\subsection{Algorithms to detect Anomalous Events from Logs}
Many of the log analysis algorithms have focused on the key area of detecting anomalous events from logs. From the survey work done by He et al. [6] and Chen et al. [7], the algorithms are either supervised or unsupervised machine learning algorithms. These algorithms may be based on classical machine learning algorithms or deep learning algorithms. Such algorithms are constrained by the availability of anomalous data with labels within the training datasets. Additionally, even with the availability of anomalous data within the log datasets, the class imbalance could pose a significant challenge with the training of the model. Even when anomalous datapoints become available from an occurrence of an incident or knowledge of how a system fault may occur or an attack vector against the system, these models would need to undergo retraining or be reconstructed to internalize the new knowledge. Hence, current algorithmic constructs for log analysis lack a form of continuous learning needed to deal with new anomalies. Our work deals with this need for model updating. 

\section{Model}
Our model uses a similar Transformer encoder [12] as part model construct to perform feature learning on log entry patterns. The Transformer encoder is capped with a fully connected multi-layer perceptron classifier with a Softmax activation layer to classify log entries as normal or anomalous. The Transformer model is trained with only normal log entries with log augmentation using log entry perturbation. As inputs to the model, character vectorization encoding is used. This creates an end-to-end model construct that is simple to use and adapt for any log source model training unlike many of other algorithms developed mentioned in our previous section for log analysis to detect anomalies requiring the use of multi-stage log processing pipeline. 

The reinforcement learning construct is applied over the Transformer model to enable the update of the model with labelled log entries.

\subsection{Character Encoding (with no parser)}\label{AA}
The character encoding is used to convert raw log entries into vectorized arrays. Aside from that, there is no other forms of preprocessing involved. Hence, raw log entries from a variety of sources could quickly be ingested into our model for training and inferences easing adoption.   
\subsection{Transformer Log Encoder with Classifier}
The model construct first applies two operations to the inputs. The first is token vectorization (word embeddings) and the other is positional encoding. This is then fed into the Transformer model [17] that is used as the primary feature learner of log entries. Our Transformer model construct comprises of two stacked layers of encoders with no decoders. The encoders have identical structures of multi-head attention layer (Equation 2) that parallelizes attention layers (Equation 1) and feed forward neural networks (Equation 3).

\begin{figure}[h]
    \centering
    \includegraphics[width=90mm,keepaspectratio]{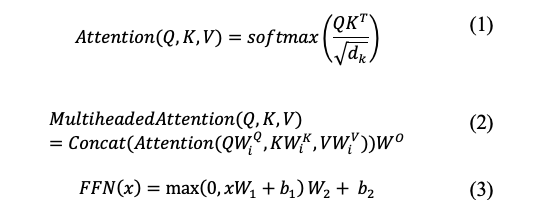}
    
\end{figure}

The model construct is finally capped with a fully connected multi-layer preceptors with a Softmax Activation layer to perform classification of the embedding vectors exiting from the Transformer encoder. 

The training objective for the Transformer encoder is to learn the features of the log entries while the objective for classifier is to accurately recognize the normal log entries from anomalies. Hence, our model would be trained with normal log entries with a balanced set of log perturbations from log normal entries using textual augmentation [18].

\subsection{Reinforcement Learning}

The Reinforcement Learning is applied to the model to update it with labelled log entry samples containing both anomalous and normal log entries. Such labelled samples could originate from human expert’s label classification during log analysis. However, to retain the model’s inference inclination towards normal log entries and to detect unusual log entries, the update training dataset would also contain log entry perturbations used in the original training of the model. Hence, the update training dataset would have a balanced of three main category of log entries. The first category is for normal, the other for labelled anomalous and the last for perturbation of the normal. Depending on the availability of the labelled anomalous log entries, the perturbed log entries would need to be generated to achieve the balanced dataset. 

The Reinforcement Learning algorithm broadly uses a goal based agent learning approach. The goal is to minimize error of the classification of log entry against the labels gathered from human analysis. This provides a form of human feedback to model for its update. The following is the pseudo code to reinforcement learning component to our model construct to update the model with acquired labels.

\begin{figure}[h]
    \centering
    \includegraphics[width=90mm,keepaspectratio]{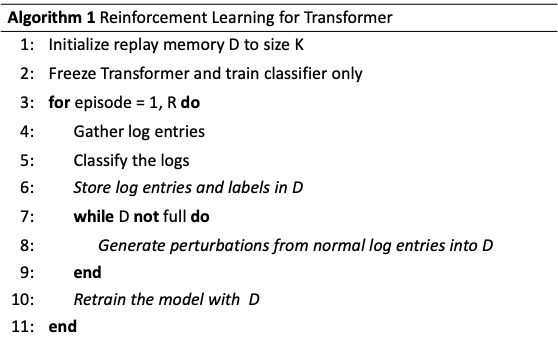}
    Pseudo Code:  Simple Reinforcement Learning Algorithm
\end{figure}

During this learning update, only the classifier is trained while the Transformer’s weights are frozen as the latter’s earlier trained self-supervised feature learning is still relevant.

\section{METHODOLOGY AND ANALYSIS}
In our experiment setup, we designed our experiment to align with those used by other past research work to facilitate the evaluation of the effectiveness of our model. We hence chose the log datasets and evaluation metrics to match those used by others accordingly.

\subsection{Log Datasets}\label{AA}
For our log datasets, we used BGL and Thunderbird [9]. These are two popular datasets typically used by researchers to evaluate their log models [7] [12]. Both log datasets are labelled.

The BGL and Thunderbird are open real-world datasets from HPC from a BlueGene/L supercomputer and Thunderbird supercomputer at Lawrence Livermore National Labs. Both share an important characteristic associated with their appearance of many new log messages in the timeline of the data, that is, the systems change over time. The following table summarizes the statistics for the two log datasets.

\begin{figure}[h]
    \centering
    \includegraphics[width=90mm,keepaspectratio]{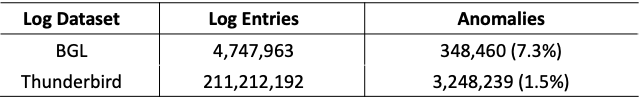}
Table 1. Log Datasets
\end{figure}

\subsection{Evaluation Metrics}
As the dataset used had binary classification labels and other research work developed their models to generate binary classification, our model was constructed with the same configuration. Additionally, most of the anomaly classification are binary that whether the log entry is normal or anomalous. We hence used precision to measure the accuracy of the model against type I error (true positive) and recall to measure the accuracy of the models against type II error (true negative). Finally, we used F1 score to measure the harmonic mean of precision and recall.

\begin{figure}[h]
    \centering
    \includegraphics[width=90mm,keepaspectratio]{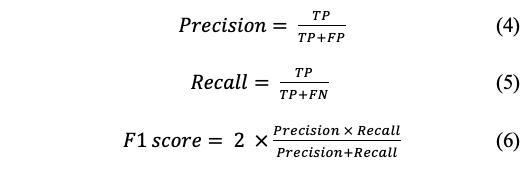}
\end{figure}

TP (True Postitive) represents the number of correctly classified anomalies against its respective labels and FP (False Positive) is the number incorrect anomaly classification. FN (False Negative) is the number of incorrect classifications of log entries as normal while the label states overwise.

\subsection{Model Preparation and Evaluation}
For model preparation and evaluation, we divided our datasets (BGL and Thunderbird) into 60:40 and 80:20 ratios for training and testing respectively. To test the training update using reinforcement learning, we created another training and test variation by further dividing the training set for 80:20 ratio into 60\% and 20\% portions with the latter portion reserved for reinforcement learning. Hence, this portion had 60:20:20. Due to resource constraints, we reduced the test set for Thunderbird dataset to create a balanced sets of normal and anomalous labels with all anomalous labelled log entries included. 

The training of the model is first done with only the normal log entries (with no anomalous labelled entries included). During training, the model is given perturbated log entries from normal ones. The proportion of the number of normal and perturbed log entries are balanced. The perturbation involved the use of a combination of word and character augmentation like randomized insertion, substitution, swapping and deletion [18]. To ensure that the Transformer learns the log entry features, the percentage of the perturbation is kept low ranging from 0.5\% to 20\%. For the experiment scenario with reinforcement learning training, the designated dataset with both normal and anomalies labels exposed.

For completeness, we compared the results of our model construct (with and without reinforcement learning) with other research works [7] using 80:20 training to test dataset ratio.

\subsection{Results and Analysis}
From our experiment testing with BGL and Thunderbird datasets, the model improves after reinforcement learning is applied.

\begin{figure}[h]
    \centering
    \includegraphics[width=90mm,keepaspectratio]{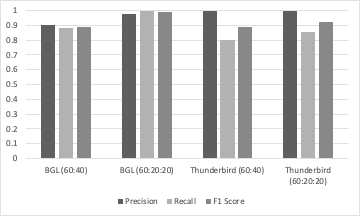}
Chart 1. Evaluation performance comparison with other algorithmic constructs with our model (with and without Reinforcement Learning)
\end{figure}

The following tables details the test results of our model in comparison with other leading log analysis algorithms and the accuracy performance metrics for the varied sample size.

\begin{figure}[h]
    \centering
    \includegraphics[width=90mm,keepaspectratio]{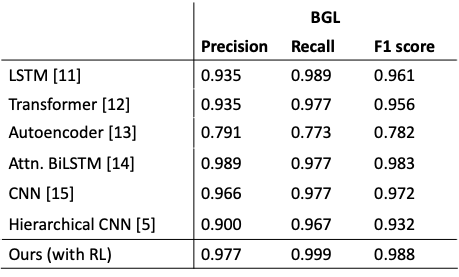}
Table 2. Evaluation performance comparison with other algorithmic constructs with our model (with and without Reinforcement Learning)
\end{figure}

From the above table, our model faired comparatively well with the BGL log dataset against other algorithmic constructs. The model improved with further updates were given with labelled samples.

\section{CONCLUSION AND FUTURE DIRECTIONS}
Our model handles the entire log analysis pipeline with one model construct. It can be trained with only normal log entries with self-induced perturbation. With reinforcement learning, the trained model can be further updated with improved performance. Hence, our approach addresses the real-world need to custom develop and deploy log analysis models for heterogenous systems without the constraints of requiring extensive log preprocessing, without the need for labels to train the model and enabling post deployment update when labels become available. 

While the model performed comparatively well with other algorithms, we will continue to improve its construct to improve its accuracy performance. One approach is to use a pretrained Transformer model with varying hyper-parameters. We intend to further test our model construct on real-world systems to support digital forensics and system resiliency.

\section*{References}

\noindent[1]	A. Pecchia, D. Cotroneo, Z. Kalbarczyk, and R.K. Iyer, “Improving log-based field failure data analysis of multi-node computing systems”, DSN’11: Proc. of the 41st IEEE/IFIP International Conference on Dependable Systems and Networks, pages 97–108. IEEE, 2011.

\noindent[2]	W. Xu, L. Huang, A. Fox, D. Patterson, and M.I. Jordon, “Detecting large-scale system problems by mining console logs”, SOSP’09: Proc. of the ACM Symposium on Operating Systems Principles, 2009.

\noindent[3]	A. Brandao and P. Georgieva, "Log Files Analysis For Network Intrusion Detection," 2020 IEEE 10th International Conference on Intelligent Systems (IS), 2020, pp. 328-333, doi: 10.1109/IS48319.2020.9199976.

\noindent[4]	M. Moh, S. Pininti, S. Doddapaneni and T. Moh, "Detecting Web Attacks Using Multi-stage Log Analysis," 2016 IEEE 6th International Conference on Advanced Computing (IACC), 2016, pp. 733-738, doi: 10.1109/IACC.2016.141.

\noindent[5]	S. Hashemi and M. Mäntylä, “OneLog: Towards End-to-End Training in Software Log Anomaly Detection”, arXiv, arXiv:2104.07324, https://doi.org/10.48550/arXiv.2104.07324.

\noindent[6]	S. He, J. Zhu, P. He and M. R. Lyu, "Experience Report: System Log Analysis for Anomaly Detection," 2016 IEEE 27th International Symposium on Software Reliability Engineering (ISSRE), 2016, pp. 207-218, doi: 10.1109/ISSRE.2016.21.

\noindent[7]	Z. Chen, J. Liu, W. Gu, Y. Su, and M. R. Lyu, “Experience Report: Deep Learning-based System Log Analysis for Anomaly Detection,”, arXiv, arXiv:2107.05908, https://doi.org/10.48550/arXiv.2107.05908.

\noindent[8]	J. Snell, K. Swersky, and R. S. Zemel, “Prototypical networks for few-shot learning”, Neural Information Processing Systems, 2017. 

\noindent[9]	A. Oliner and J. Stearley, “What supercomputers say: A study of five system logs”, 37th Annual IEEE/IFIP International Conference on Dependable Systems and Networks (DSN’07). IEEE, pp 575–584, 2007.

\noindent[10]	V. H. Le and H. Zhang, “Log-based Anomaly Detection without Log Parsing”, 2021 36th IEEE/ACM International Conference on Automated Software Engineering (ASE), Nov 2021.

\noindent[11]	M. Du, F. Li, G. Zheng, and V. Srikumar, “Deeplog: Anomaly detection and diagnosis from system logs through deep learning”, Proceedings of the 2017 ACM SIGSAC Conference on Computer and Communications Security, 1285–1298, 2017.

\noindent[12]	S. Nedelkoski, J. Bogatinovski, A. Acker, J. Cardoso, and O. Kao, “Self-attentive classification-based anomaly detection in unstructured logs”, 2020 IEEE International Conference on Data Mining (ICDM), Nov 2020.

\noindent[13]	A. Farzad and T. A. Gulliver, “Unsupervised log message anomaly detection”, ICT Express 6, 3, 229–237, 2020.

\noindent[14]	X. Zhang, Y. Xu, Q. Lin, B. Qiao, H. Zhang, Y. Dang, C. Xie, X. Yang, Q. Cheng, Z. Li, et al, “Robust log-based anomaly detection on unstable log data”, Proceedings of the 2019 27th ACM Joint Meeting on European Software Engineering Conference and Symposium on the Foundations of Software Engineering, 807–817, 2019.

\noindent[15]	S. Lu, X. Wei, Y. Li, and L. Wang, “Detecting anomaly in big data system logs using convolutional neural network”, 2018 IEEE 16th Intl Conf on Dependable, Autonomic and Secure Computing, 16th Intl Conf on Pervasive Intelligence and Computing, 4th Intl Conf on Big Data Intelligence and Computing and Cyber Science and Technology Congress (DASC/PiCom/DataCom/CyberSciTech), IEEE, 151–158, 2018.

\noindent[16]	V. H. Le and H. Zhang, “Log-based Anomaly Detection without Log Parsing”, 2021 36th IEEE/ACM International Conference on Automated Software Engineering (ASE), Nov 2021.

\noindent[17]	A.Vaswani, N. Shazeer, N. Parmar, J. Uszkoreit, L. Jones, A.N. Gomez, Ł. Kaiser, and I. Polosukhin, “Attention is all you need,” in Advances in neural information processing systems, 2017, pp. 5998–6008.

\noindent[18]	E. Ma, NLP Augmentation, https://github.com/makcedward/nlpaug, 2019.

\end{document}